\documentclass[conference]{IEEEtran}
\IEEEoverridecommandlockouts
\usepackage{cite}
\usepackage{amsmath,amssymb,amsfonts}
\usepackage{algorithmic}
\usepackage{graphicx}
\usepackage{textcomp}
\usepackage{multirow}
\usepackage{subcaption}
\usepackage{graphicx}
\usepackage{float}
\usepackage{color,soul}
\usepackage{enumitem}

\usepackage{booktabs}
\usepackage{pifont}
\usepackage{soul}
\usepackage{color}
\usepackage[usenames,dvipsnames]{xcolor}

\definecolor{peach-orange}{rgb}{1.0, 0.8, 0.6}
\definecolor{pastelyellow}{rgb}{0.99, 0.99, 0.59}
\newcommand{\xmark}{\ding{55}}%

\def\BibTeX{{\rm B\kern-.05em{\sc i\kern-.025em b}\kern-.08em
    T\kern-.1667em\lower.7ex\hbox{E}\kern-.125emX}}
\begin{document}

\title{Optimized preprocessing and Tiny ML for Attention State Classification\\

}

\author{\IEEEauthorblockN{\textsuperscript{} Yinghao Wang, Rémi Nahon, Enzo Tartaglione, Pavlo Mozharovskyi, and Van-Tam Nguyen}
\IEEEauthorblockA{\textit{LTCI, Télécom Paris, Institut Polytechnique de Paris} \\
yinghao.wang@telecom-paris.fr}

}

\maketitle

\begin{abstract}

In this paper, we present a new approach to mental state classification from EEG signals by combining signal processing techniques and machine learning (ML) algorithms. We evaluate the performance of the proposed method on a dataset of EEG recordings collected during a cognitive load task and compared it to other state-of-the-art methods. The results show that the proposed method achieves high accuracy in classifying mental states and outperforms state-of-the-art methods in terms of classification accuracy and computational efficiency.
\end{abstract}

\begin{IEEEkeywords}
Tiny ML, Attention level detection, EEG, Brain-computer interface
\end{IEEEkeywords}

\section{Introduction}

Attention is a cognitive process that involves the selection and prioritization of sensory information to be processed~\cite{posner1994attention}. The ability to accurately classify attention states using electroencephalography (EEG) could have important implications for a wide range of applications, from monitoring cognitive load in educational settings~\cite{liu2013recognizing} to developing neurofeedback systems to improve cognitive performance~\cite{lim2012brain}\cite{enriquez2017eeg}.\\
\noindent Due to the non-stationary nature of EEG signals, as well as their time-varying spectral characteristics~\cite{boashash2015time}, it is necessary to analyze EEG signals in the joint time-frequency domain rather than the time-only or the frequency-only domain only\cite{oppenheim1999discrete}. In recent research, machine learning-based classifiers have shown high potential for distinguishing mental states using spectrogram-based features. In \cite{aci2019distinguishing}, the Support Vector Machine (SVM) was used to classify three mental states (focused, unfocused, and drowsy) with 91.72\% accuracy when training and testing were performed on a single subject with seven channels. In \cite{yang2018toward}, an accuracy of 96.11\% was achieved using an SVM for drowsiness detection on three-channel data from a single subject. Previous work also shown that, in some cases, deep learning-based algorithm is more efficient than conventional machine learning in this type of task~\cite{thodoroff2016learning}\cite{gao2021complex}\cite{zhu2021vehicle}, which motivates us to optimize the preprocessing process to make it more suitable for Tiny ML.\\
\noindent In this work, we explore the effect of different compositions of labeled EEG data on the classification results. In particular, for feature extraction, we conduct a quantitative study on the parameters of the spectrogram-based features extraction, and for the classification algorithm, we implement Support Vector Machines (SVM), eXtreme Gradient Boosting (XGB), and Random Forest (RF) along with two neural network models.  
\section{Material and method}
\label{sec:material}
\begin{figure*}[htbp]

\centerline{\includegraphics[width=.98\textwidth]{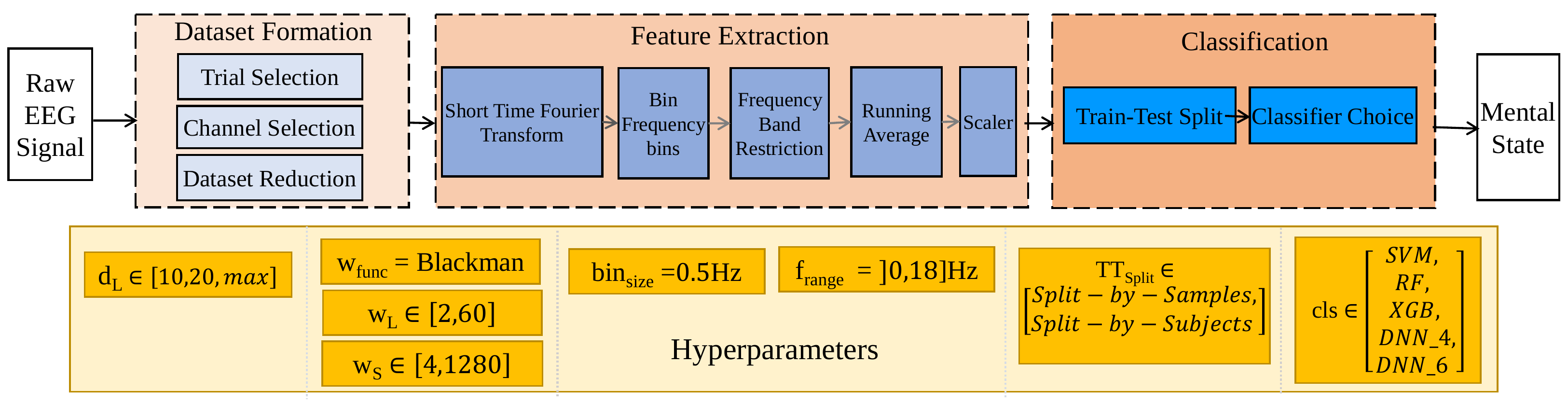}}
\caption{Representation of the whole process undertaken from the raw signal to the mental state detection.}
\label{hyperparams}
\end{figure*}

Fig.~\ref{hyperparams} provides an overview of the approach undertaken in this work to go from raw EEG signal to the final mental state classification and the main hyperparameters involved. We use the dataset presented in Sec.~\ref{sec:dataset} proposed by \cite{aci2019distinguishing}. First, we proceed to a pre-selection of the data (Sec.~\ref{sec:dataset_formation}). Then we perform our feature extraction (Sec.~\ref{sec:feature_extraction}) that will improve the performance of the different classifiers we use for the mental classification task (Sec.~\ref{sec:classifiers}).

\subsection{Raw dataset used}
\label{sec:dataset}
In this work, we use a publicly available dataset collected from five participants by  \cite{aci2019distinguishing}. 25 hours of EEG recordings were made, and each trial lasted 35 to 55 minutes. The three studied mental states are the following:  
\begin{enumerate}[noitemsep,nolistsep]
    \item \emph{Focused} but passive attention,
    \item \emph{Unfocused} or detached but awake,
    \item \emph{Drowsy} or on the verge of falling asleep.
\end{enumerate} 
\noindent The task consists of controlling a computer-simulated train using the program “Microsoft Train Simulator”. During the first 10 minutes of each trial, participants are engaged in focused control of the simulated train, paying close attention to the simulator controls. For the next 10 minutes, the train moves on a flat road without any control from the participant, who entered a de-focused state. For the remainder of the experiment, participants are allowed to relax freely, close their eyes, and doze off, as they wish. The supervisor determines the end time of the experiment when the participant completes the mental drowsiness state.\\
\noindent This dataset provides EEG signals from 7 channels at a sampling rate of 128Hz, which are identified as F3, F4, Fz, C3, C4, Cz, and Pz in the standard 10-20 electrode system~\cite{klem1999ten}. More detailed descriptions of the experimental paradigm were presented in~\cite{aci2019distinguishing}.
\subsection{Dataset formation}
\label{sec:method}
\label{sec:dataset_formation}
\noindent\textbf{Trial selection.} Various factors such as motion artifacts or unfamiliarity with the experimental procedures can lead to mislabeling of the collected data. To minimize these risks, the choice made in \cite{aci2019distinguishing} is strictly adhered to and the first two trials of each participant from the final dataset are deleted.\\
\noindent\textbf{Dataset length reduction.}
Algorithmic fairness plays an important role in ML and imposing fairness constraints during learning is a common approach. Imbalanced datasets lead to a lack of generalizability not only of classification but also of fairness properties, especially in over-parameterized models \cite{deng2022fifa}. To address this problem, we propose three different paradigms regarding the value of $d_{L}$, the length (in minutes) of the drowsy state portion of the trial that we retain:
\begin{itemize}[noitemsep,nolistsep]
    \item $d_{L}\!=\!10$: the dataset is balanced, with ten minutes in each state for each trial, but the number of samples used for training is significantly reduced,    
    \item $d_{L}\!=\!20$: the dataset is balanced between drowsiness and non-drowsiness data and the duration of each trial is reduced to 40 minutes,
    \item $d_{L}\!=\!max$: the dataset is the original unbalanced one.
\end{itemize}
\noindent\textbf{Channel selection.} EEG headsets typically have dozens or even hundreds of electrodes that capture electrical signals from the brain. While using fewer electrodes can make the headset less cumbersome and more convenient in practice, it can also limit the amount and quality of data collected, which can affect the accuracy of the analysis. We performed most of our analysis with 7 electrodes, as in \cite{aci2019distinguishing}. From the perspective of a wearable device, we examined how decreasing the number of EEG channels affects classification performance.
\subsection{Feature extraction}
\label{sec:feature_extraction}
\noindent\textbf{STFT window length.} We chose to build our feature vectors from spectrograms obtained by Short Term Fourier Transform (STFT). The choice of the STFT window length $w_{L}$ is a trade-off between frequency resolution and temporal resolution. We explore $w_{L} \in [2, 60]$ (in seconds) because we want to achieve a frequency resolution of 0.5Hz (equivalent to $w_{L} = 2$) and we intend to use our method in a real-time perspective which makes windows bigger than 60 seconds intractable.\\
\noindent\textbf{STFT window shift.} 
Inspired by \cite{mu2017driver}, we assume that the overlap between windows also has an impact on the efficiency of the feature extraction process. In particular, a small window shift ${w_{S}}$ provides more samples, resulting in more training data, thus improving ML performance. However, it is also known that too much dependence between the input samples leads to overfitting and loss of generalization of the models. We, therefore, perform a joint optimization of $w_{S}$ and $w_{L}$, because the overlap can be calculated from them. We consider ${w_{S} \in [4,1280]}$, ie. from the minimum accessible in terms of memory to ten seconds of shift between data points, which reduces too much the number of points for classification.\\
\noindent\textbf{STFT window function.} EEG signals often contain large spurious signals near the frequency of interest, so the \emph{Blackman} function is used as in \cite{aci2019distinguishing},
because the resulting spectrum has a wide peak, but good side-lobe compression~\cite{bekka2003effect}.\\
\noindent\textbf{Frequency binning.} The setting in \cite{aci2019distinguishing} is used to bin the frequency bands using $bin_{size}=0.5$ (in Hz), which can lead to higher accuracy while also reducing the ML training time.\\
\noindent\textbf{Band restriction.} We restricted the frequencies to a range $f_{range} = (0,18]$ (in Hz) as it is known that the significant changes in EEG signals occur within that range \cite{alazrai2022deep}.\\
\noindent\textbf{Running average.} The spectrograms are smoothed over time using a 15-second moving average to reduce noise. The final feature vector is then formed by converting the power values at each time point into decibels.\\
\noindent\textbf{Scaling.} To balance the influence of each feature on classification, regardless of the range of values it takes \cite{bengio2003no}, we apply {$scaler = standardization$} to the feature vector. \\
\noindent\textbf{Train-test split.} We use two distinct methods to divide the dataset into training and testing sets to fulfill different goals:\\
The \emph{Split-by-Samples} paradigm reflects the ability of the model to apply what it learns to the individual it has been training on. We divide it into two sub-paradigms regarding the number of subjects involved. With \emph{Subject-Specific}, as we train and test on a single subject, we can efficiently identify changes in an individual's brain activity in response to various stimuli or tasks. With \emph{Common-Subject}, we train and test on all subjects at once, which is widely used in the literature as it makes it possible to train with more variability in the data.\\
One of the major challenges we face in EEG classification lies in the generalization of what is learned from one set of individuals to another one, as the features picked up by the models tend to vary considerably from subject to subject. To test our method on this point, we propose the \emph{Split-by-subject} paradigm that involves training and testing on different individuals. Its application in this paper referred to as \emph{Leave-One-Out}, consists of training the classifier on four subjects and testing on the fifth.
\subsection{Classifiers used}\label{sec:classifier}
\label{sec:classifiers}
In this work, we use four different families of algorithms to perform our EEG classification tasks:\\
\noindent\textbf{RF} is efficient to process noisy and unbalanced data such as those encountered in EEG signals~\cite{edla2018classification}.\\
\noindent\textbf{SVM} is used for its ability to handle high-dimensional data and its efficacy on both linear and non-linear problems ~\cite{zhang2021support}.\\
\noindent\textbf{XGB} is known for its speed, accuracy, and ability to handle large and high-dimensional datasets~\cite{tiwari2019multiclass}. \\
\noindent\textbf{Neural Networks} (NN) are effective and popular for EEG classification tasks, especially when combined with appropriate feature extraction and preprocessing methods \cite{thodoroff2016learning}. We use two neural network architectures in this paper. \emph{DNN\_4} is a simple multilayer perceptron of four layers (with two hidden 64-neuron layers). We also use a deeper network, \emph{DNN\_6}, that consists of \emph{DNN\_4} with two supplementary internal layers (of 128 neurons) and two dropouts layers before the last hidden layer to address the increased risk of overfitting known to be inherent to larger neural networks \cite{zhang2021understanding}.
\section{Experiments and results}
\label{sec:experiments}
\subsection{Data length study}\label{sec:downsampling}
\begin{figure}[htbp]
\centerline{\includegraphics[width=0.49\textwidth,height=0.36\textwidth]{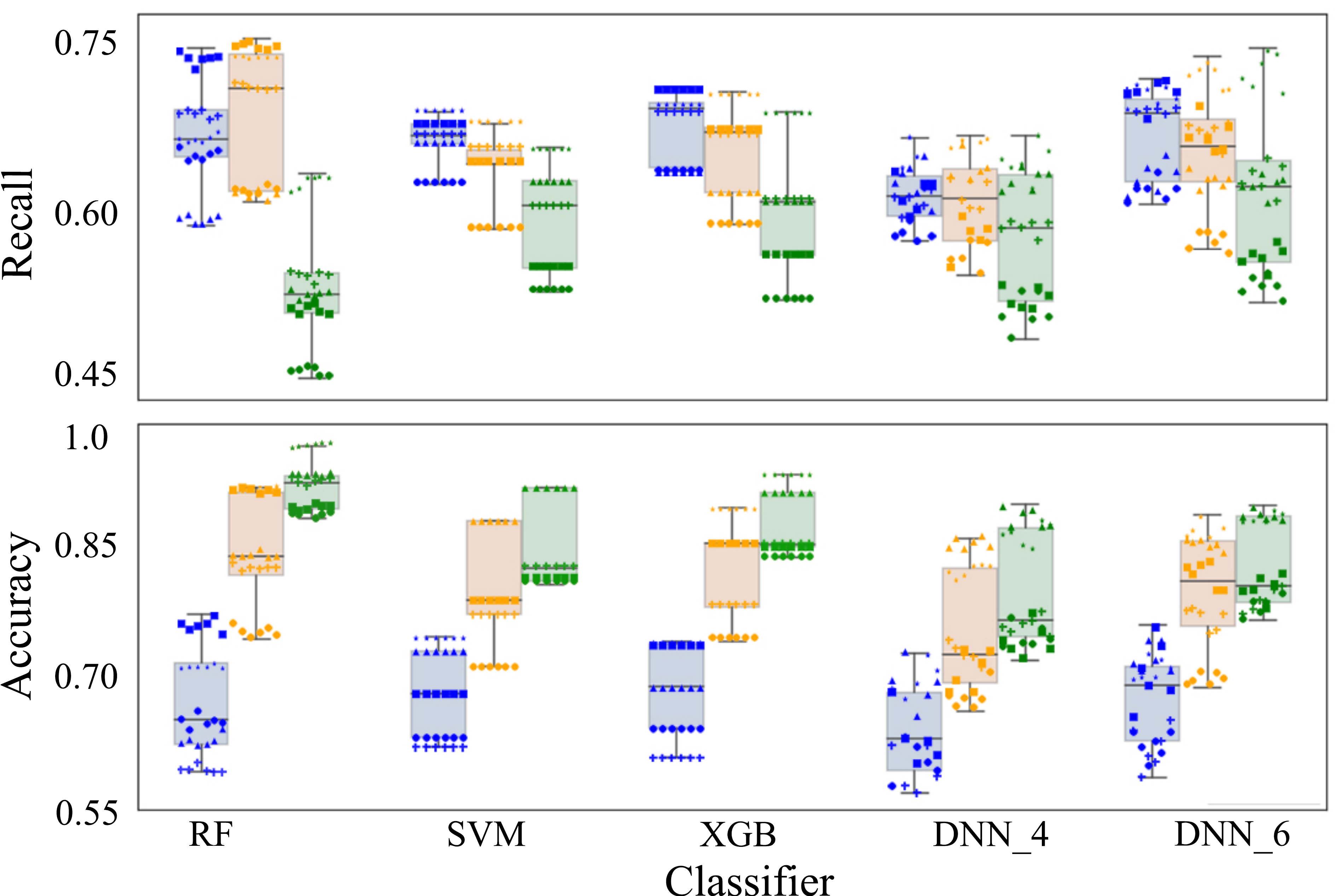}}
\caption{Accuracy (top part) and recall on the ''drowsy'' state (bottom part) are evaluated for 5 classifiers for different $d_L$ values. The results for all five subjects are represented by different shapes of points, arranged according to the order of the 6 random seeds used. }
\label{fig4}
\end{figure}

In this section, we use the seven channels of EEG signal provided by the dataset, while keeping the hyperparameters introduced in Sec.~\ref{sec:method} as presented in~\cite{aci2019distinguishing}, without any modification. The \emph{Leave-One-Out} paradigm is used here as it is the most demanding. Fig.~\ref{fig4} shows the results of the five classifiers presented in Sec.~\ref{sec:classifier}. In addition, we chose to use a balanced accuracy instead of the traditional one that would be biased towards the more represented ''drowsed'' class.\\
\noindent\textbf{Accuracy.}
As shown in the top part of Fig.~\ref{fig4}, the classification accuracy is higher for \textbf{$d_{L} = 20$} than for \textbf{$d_{L} = max$}, across the five subjects. However when comparing the paradigms \textbf{$d_{L} = 20$} and \textbf{$d_{L} = 10$}, the performances are very close.\\
\noindent\textbf{Recall.} Regarding the recall (in the bottom part of Fig.~\ref{fig4}), the best performances are obtained for 
 \textbf{$d_{L} = max$}, and the results for \textbf{$d_{L} = 20$} are very close to it. On the other hand, the results for \textbf{$d_{L} = 10$} are much worse. \\
\noindent These results show that correctly balancing data is useful for ML training. However, finding the full balance of the dataset may significantly reduce the training samples, which will then hurt the training. As it seems to always yield good results, we use the ${d_{L} = 20}$ paradigm in all subsequent experiments. 

\subsection{STFT quantitative study}\label{sec:stft}

\begin{figure}[htbp]
\centerline{\includegraphics[width=0.49\textwidth]{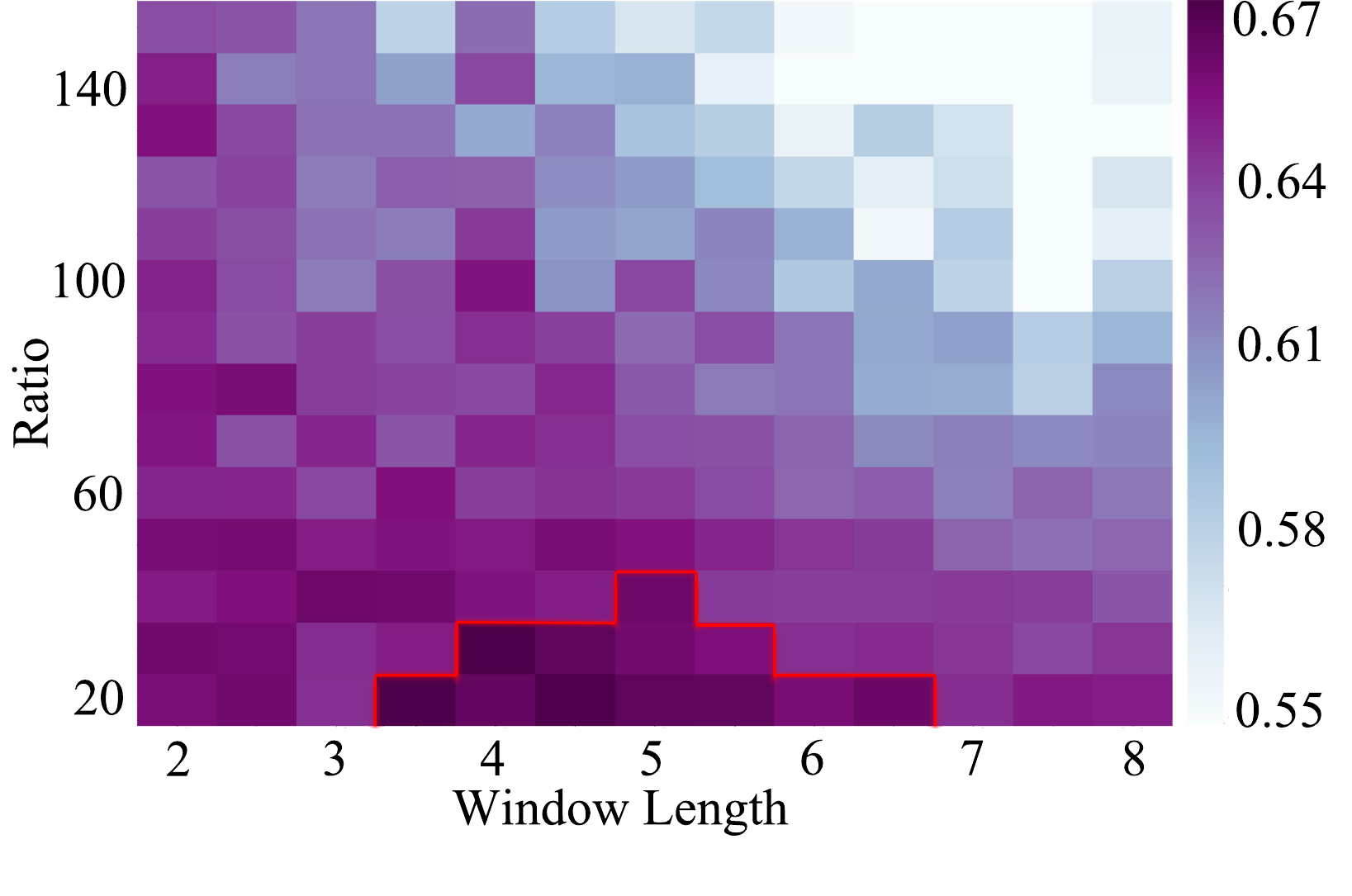}}
\caption{Accuracy on the \emph{Leave-One-Out} paradigm for different feature extraction configuration: Variation of the window length (in seconds) and of the ratio $\frac{w_S}{w_L}$. The red lining frames the best combinations.}
\label{fig6}
\end{figure}

In order to study the effect of different STFT parameters on the performance of the classifier, the \emph{Leave-One-Out} paradigm is chosen, due to the generalization perspective, but also due to the higher validity of the extracted features. Since large NN models contain huge amounts of parameters, the DNN\_6 classifier is very sensitive to feature quality. \\
\noindent As shown in Fig.~\ref{fig6} the feature quality deteriorates with increasing window length and shift scale. In particular, we obtain the best features in the red triangular shape region. \\
\noindent In order to study this area more thoroughly, we use sample-sliding instead of a window length ratio to avoid a decrease in accuracy caused by a lack of training data. In this way, we decouple the relationship between window length and the number of data points. As shown in Fig.~\ref{fig6}, the best result is obtained when ${w_{S} = 128}$. The accuracy decreases as {$w_{L}$} increases, and the best result is obtained for ${w_{L} = 4}$.\\
\noindent Generally, different combinations of \textbf{$w_{L}$} and \textbf{$w_{S}$} affect the extracted features in two ways: the independence between the samples used to train the ML algorithm, and the number of samples. In the ideal case, we want each feature to provide unique information to the algorithm, and not have redundant or overlapping information. However, this requires a large amount of labeled data, and the cost of the data acquisition is a pressing problem for Tiny ML. In this situation, data enhancement is achieved by adding some overlap between samples, which improves accuracy by 6\%. 

\begin{figure}[htbp]
\centerline{\includegraphics[width=0.49\textwidth]{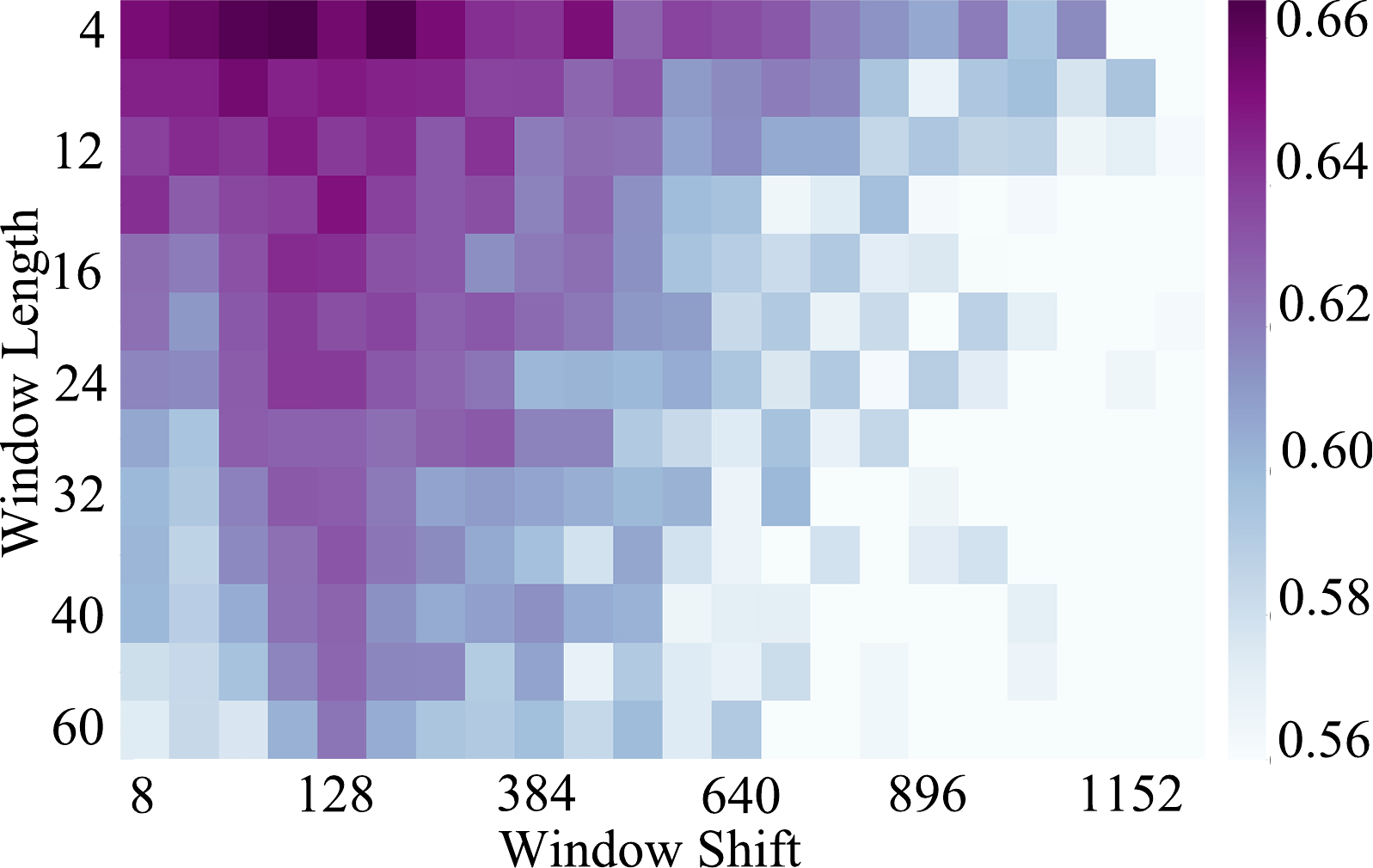}}
\caption{Accuracy on the \emph{Leave-One-Out} paradigm for different feature extraction configuration: Variation of the window length (in seconds) and of window shift (in samples).}
\label{fig8}
\end{figure}
\subsection{Results of classification}
\begin{figure}[htbp]
\centerline{\includegraphics[width=0.47\textwidth]{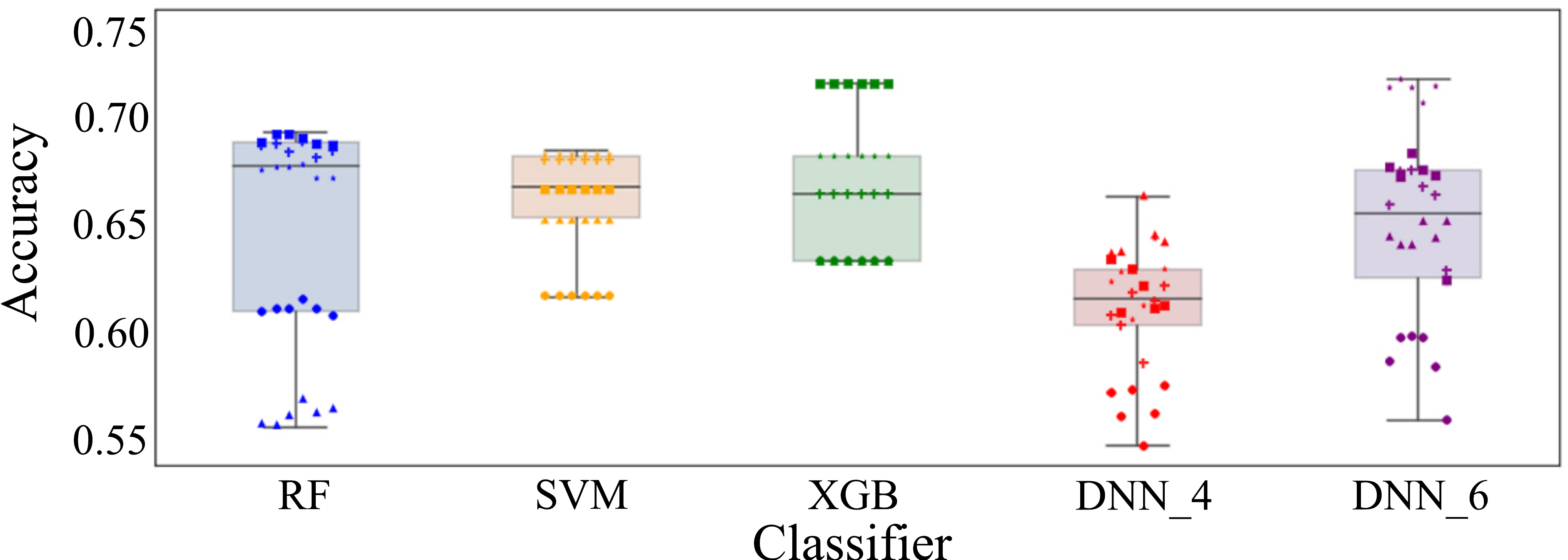}}
\caption{Accuracy on the \emph{Leave-One-Out} paradigm for five classifiers. As in Fig~\ref{fig4}, the results for each individual subject and seed are represented by specific points.}
\label{fig7}
\end{figure}

In this section, we evaluate the classification performance on the three train-test split paradigms. In each case, we use the optimal hyperparameters found in Sec.~\ref{sec:experiments}.\\
\begin{table}[]
\centering
\caption{Results on the \emph{Subject-Specific} paradigm ($\dagger$ indicates the results taken from \cite{aci2019distinguishing}).}
\begin{tabular}{ccccc}
\toprule
\multirow{2}{*}{\bf Classifier}                                & \bf Best  & \bf Average         &\bf Feature         &\bf Fine                 \\ 
                                                            & \bf accuracy &\bf accuracy   & \bf engineering &\bf tuning\\\midrule
RF                  &  99.6\%   & 99.3\%          & \checkmark & \checkmark \\ 
XGB                 & \underline{99.8\%}   & \underline{99.6\%}          & \checkmark & \checkmark \\ 
SVM                 & \textbf{99.9\%}   & \textbf{99.8\%} & \checkmark & \checkmark \\ 
DNN\_4              & \underline{99.8\%}   & \underline{99.6\%}          & \checkmark & \checkmark \\ 
DNN\_6              & \textbf{99.9\%}   & \underline{99.6\%}          & \checkmark & \checkmark \\ 
SVM & 96.7\%$^\dagger$   & 91.7\%$^\dagger$          & \checkmark &  \xmark   \\ \bottomrule                   
\end{tabular}%
\label{channels_ablation}
\end{table}
\noindent\textbf{Feature engineering effects.} To test the effects of our approach, 
TABLE~\ref{channels_ablation}, performed in the \emph{Subject-Specific} paradigm, shows that the detection of attentional states can be achieved effectively using handmade features. Compared to the results obtained by \cite{aci2019distinguishing}, with the same classifier (SVM), our feature extraction with fine-tuned parameters significantly improves the yielded accuracy (from 96.7\% to 99.9\% for the best accuracy and from 91.7\% to 99.8\% for the average accuracy). This shows the efficacy of our feature extraction method.\\
\noindent\textbf{Channel selection.}\label{sec:cs} We investigate the effect of the number of EEG electrodes used to acquire the signal on the prediction results. We perform this study in \emph{Common-Subject} paradigm as it helps train on the most data. From the ablation study of the channels, we know that the three positions that have the greatest influence on classification accuracy are Fz, F3, and Pz. TABLE.~\ref{t2} shows the accuracy obtained for the best channel combination for multiple numbers of channels. 
\begin{table}[]
\centering
\caption{Accuracy on the \emph{Common-Subject} paradigm for different numbers of channels.}
\setlength\tabcolsep{14pt}%
\begin{tabular}{@{}ccccc@{}}
\toprule
\multirow{2}{*}{\bf Classifier} & \multicolumn{4}{c}{\bf Number of channels used} \\ 
 & 7 & 3 & 2 & 1 \\ \midrule
RF & 97.3\% & 96.6\% & \textbf{95.8\%} & \textbf{92.6\%} \\ 
SVM & \textbf{99.2\%} & \textbf{97.6\%} & \underline{94.0\%} & 82.8\% \\ 
XGB & 98.0\% & 95.0\% & 91.2\% & 81.9\% \\ 
DNN\_4 & \underline{98.3\%} & 96.7\% & 93.9\% & \underline{87.8\%} \\ 
DNN\_6 & 98.1\% & \underline{97.3\%} & 93.6\% & 87.4\% \\ \bottomrule
\end{tabular}%
\label{t2}
\end{table}
\noindent The RF algorithm can effectively avoid overfitting and obtain the best accuracy when the amount of data is small. This is especially evident when there is only one channel. In NN algorithms, overfitting is also mitigated by adding a dropout layer, resulting in better accuracy. The SVM algorithm performs best when the amount of data is large. 
Overall, even though we use only two electrodes to collect EEG data, we achieve more than 95\% accuracy, demonstrating the validity of our extracted features and is also promising for real-world applications.\\
\noindent \textbf{Models generalization.}\label{sec:leave-one-out}
Generalizing feature extraction for EEG signals has been an important problem for researchers due to the variation of EEG signals among individual subjects. Effective features and appropriate algorithms can help solve this problem, even in this very challenging paradigm. To test the performance of the models on that aspect, we used the \emph{Leave-One-Out} paradigm. As shown in Fig.~\ref{fig7}, we achieve the best accuracy of 71.1\% using the DNN\_6 classifier under this paradigm. In addition, the deep learning algorithm also shows good potential for more balanced performance across subjects.

\section{Conclusion}
\label{sec:conclusion}
\noindent In this paper, we presented in detail the feature engineering of EEG signals, verified the effectiveness and practicality of the extracted features through comprehensive experiments, and compared established classification algorithms. The results show that our careful choice of preprocessing parameters and their optimization significantly improves the final result compared to the state-of-the-art (over 3\% for best accuracy and over 8\% for average accuracy). This allows very high accuracy in \emph{Subject-specific} and \emph{Common-Subject} paradigms.
At the same time, in the \emph{Leave-One-Out} paradigm, the proposed algorithms are still far from the two other paradigms. This can be explained to a large extent by an insufficient number of individuals. However, the result from NN models are promising and motivates us to explore more efficient feature extraction methods and to design better tiny ML algorithms.

\newpage
{\small
\bibliographystyle{unsrt}
\bibliography{ref}
}
\end{document}